\documentclass{article}

\usepackage[T1]{fontenc}
\usepackage[utf8]{inputenc}

\usepackage{microtype}
\usepackage{graphicx}
\usepackage{overpic}
\usepackage{booktabs} % for professional tables
\usepackage{tabularx}
\usepackage{algorithm,algpseudocode}

\usepackage{hyperref}
\usepackage{fontawesome5}
\usepackage{pifont}
\usepackage{multirow}
\usepackage{enumitem}

\usepackage{nicematrix,tikz} 
\usepackage{arydshln}
\usepackage{caption,subcaption}

\usepackage{iac_pkg}

\usepackage[accepted]{icml2025}

\usepackage{amsmath}
\usepackage{amssymb}
\usepackage{mathtools}
\usepackage{amsthm}

\usepackage[capitalize]{cleveref}

\theoremstyle{plain}

\theoremstyle{definition}

\theoremstyle{remark}

\usepackage[textsize=tiny]{todonotes}

\icmltitlerunning{Inverse Language Modeling towards Robust and Grounded LLMs}

\begin{document}

\twocolumn[
\icmltitle{Inverse Language Modeling towards Robust and Grounded LLMs}

\icmlsetsymbol{equal}{*}

\begin{icmlauthorlist}
\icmlauthor{Davide Gabrielli}{equal,sapienza}
\icmlauthor{Simone Sestito}{equal,sapienza}
\icmlauthor{Iacopo Masi}{sapienza}
\end{icmlauthorlist}

\icmlaffiliation{sapienza}{\methodimagetag{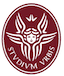}~~Sapienza University of Rome, Computer Science Department,~~ \methodimagetag{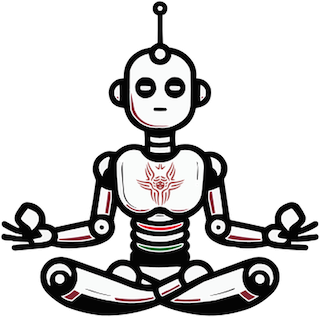} \href{https://omnai.di.uniroma1.it}{OmnAI Lab}}

\icmlcorrespondingauthor{Davide Gabrielli}{davide.gabrielli@uniroma1.it}

\icmlkeywords{Artificial Intelligence, LLMs, robustness, inverse problems}

\vskip 0.3in
]

\printAffiliationsAndNotice{\icmlEqualContribution}

\begin{center}
\centering
\textit{``A causal model looks ahead, but only its gradients disclose the pasts that might have built that future.''}
\end{center}
\vspace{1em}

\begin{abstract}
The current landscape of defensive mechanisms for LLMs is fragmented and underdeveloped, unlike prior work on classifiers.
To further promote adversarial robustness in LLMs, we propose Inverse Language Modeling (ILM), a unified framework that simultaneously 1) improves the robustness of LLMs to input perturbations, and, at the same time, 2) enables native grounding by inverting model outputs to identify potentially toxic or unsafe input triggers.
ILM transforms LLMs from static generators into analyzable and robust systems, potentially helping RED teaming.
ILM can lay the foundation for next-generation LLMs that are not only robust and grounded but also fundamentally more controllable and trustworthy.
The code is publicly available at \href{https://github.com/davegabe/pag-llm}{github.com/davegabe/pag-llm}.

\end{abstract}

\section{Introduction}
Large Language Models (LLMs) excel at natural language tasks and reasoning. Today, a single foundation model can handle a wide range of NLP tasks. However, LLMs are still prone to hallucinations and remain sensitive to input variations, such as adversarial prompts. Recent work indicates that even sensical perturbations~\cite{zou2023universal,melamed2024prompts} can trigger these issues, highlighting the potential for backdoors~\cite{carlini2024poisoning}.

\begin{figure}
\vspace{.5cm}
\begin{overpic}[width=\linewidth]{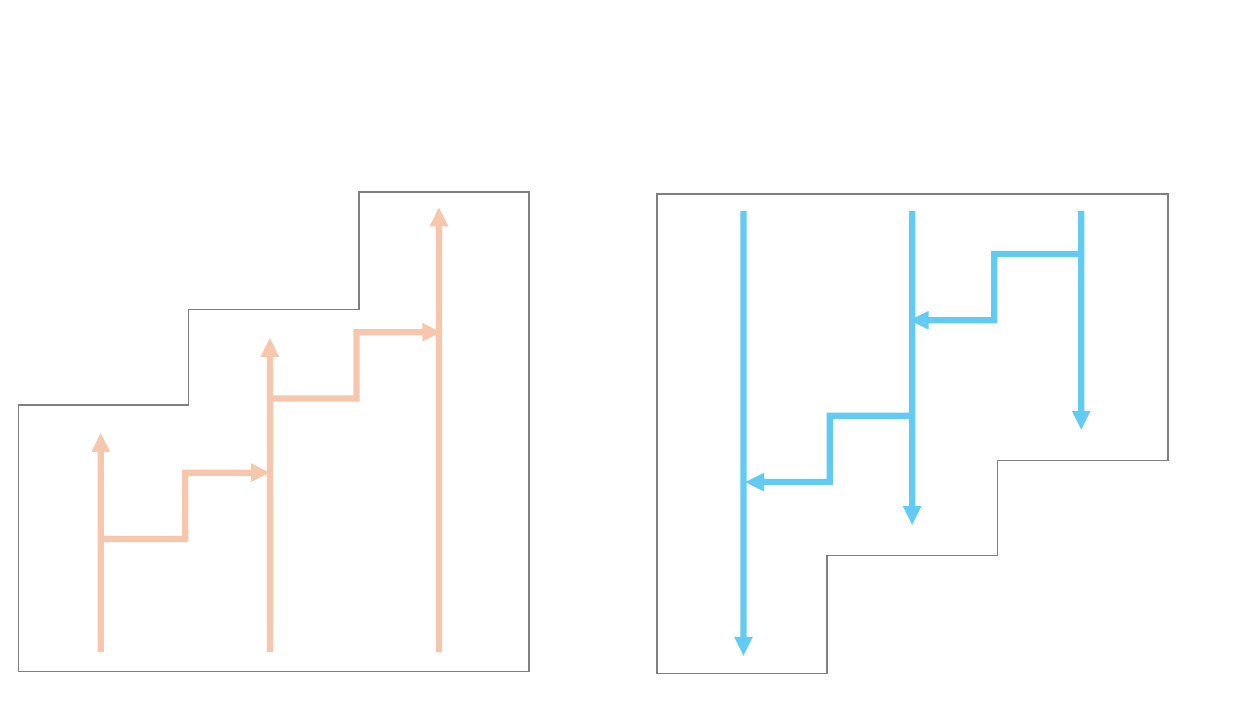}
    \put(5.5,0){$\bx_1$}\put(19.5,0){$\bx_2$}\put(33.5,0){$\bx_3$}
    \put(5.5,27){$\bx_2$}\put(19.5,34){$\bx_3$}\put(33.5,44){$\bx_4$}
    \put(60,43.5){$\bx_2$}\put(72,43.5){$\bx_3$}\put(87,43.5){$\bx_4$}
    \put(60,0){$\bx_1$}\put(72,10){$\bx_2$}\put(87,18){$\bx_3$}
    \put(17,59.5){{\footnotesize \textbf{Now}}}
    \put(68,59.5){{\footnotesize \textbf{Proposed}}}
    \put(0,55){{\footnotesize Autoregressive forward}}
    \put(55,55){{\footnotesize Autoregressive backward}}
    \put(5,49){{\scriptsize $p(\bx_i|\bx_1,\ldots,\bx_{i-1})$}}
    \put(50,49){{\scriptsize$p\big(\bx_{i-1}| \nabla_{\bx_{i-1}}p(\bx_i|\bx_1,\ldots,\bx_{i-1})\big)$}}
\end{overpic}
\caption{Illustration of Inverse Language Modeling (ILM) setup. Forward pass predicts next tokens, backward pass reconstructs inputs from gradients.\vspace{-5pt}}
\label{fig:inverse_lm_idea_schema}
\end{figure}
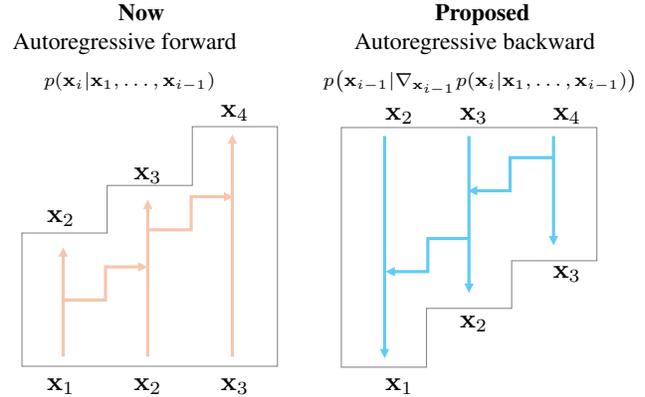

These problems emphasize the need for adversarial training (AT) tools for LLMs. Alas, the literature on this topic is not as dense as for deep classifiers, and yet the security of LLMs to adversarial perturbations remains an open challenge. Moreover, the training of LLMs is very costly, and thus, applying AT could only worsen the issue.

Efficient solutions for AT for LLMs intercept a pressing need~\cite{xhonneux2024efficient}. 
In this work, we define \textbf{robustness} as reduced sensitivity to adversarially perturbed prompts, and \textbf{grounding} as ensuring that LLMs ``know what they have been asked'', addressing evidence that they often fail to represent their own knowledge faithfully~\cite{melamed2024prompts,bender2021dangers}.

In light of this, the objectives are twofold.
\pointbone{} \tbf{Robustness}: we will study and analyze a new, fast and efficient AT method for LLMs, proposing \tbf{Inverse Language Modeling (ILM)}, which takes inspiration from years of progress on robust classifiers and leverages the notion that \tbf{Perceptually Aligned Gradients (PAG)} imply robustness~\cite{ganz23aPAG}. LLMs are trained in forward-mode, i.e., when given a text prompt $\bx$, the transformer~\cite{vaswani2017attention} predicts its completion $\by$ with self-supervision\footnote{For clarity, we denote the text prompt as $\bx$, which in practice corresponds to a sequence of input tokens ${\bx_0, \dots, \bx_{i-1}}$. The target sequence $\by$ is then the one-step left-shifted version of $\bx$.}.
ILM extends the forward modeling paradigm in a backward direction (see \autoref{fig:inverse_lm_idea_schema}). Specifically, given an output $\by$ (the answer), we ask whether the LLM can recover the conditioning prompt $\bx$.
% Importantly, ILM does not reverse the token sequence. It recovers the input prompt by performing gradient-based alignment that is informed by both the model’s output probabilities and the representations accumulated at each layer during the forward pass.
\pointbtwo{} \tbf{Grounded LLMs}: the second goal is a byproduct of \pointbone{} and allows RED teaming to better investigate what may generate a malicious output $\by$ by inverting it. 

Importantly, ILM does not reverse the token sequence. It recovers the input prompt by performing gradient-based alignment that is informed by both the model’s output probabilities and the representations accumulated at each layer during the forward pass.
\section{Prior Work}

Research on adversarial attacks against Large Language Models (LLMs) has advanced significantly, particularly in generating adversarial suffixes designed to bypass alignment safeguards. Early techniques, such as HotFlip~\cite{ebrahimi2018hotflip} and Greedy Coordinate Gradient (GCG)~\cite{zou2023universal}, focused on manipulating the input text or its embedding gradients to elicit undesired behavior from LLMs. GCG modifies token selections iteratively based on gradient information. Subsequent enhancements, including Probe Sampling~\cite{zhao2024accelerating} and token similarity-based heuristics~\cite{li2024faster}, have improved its efficiency.

More recent methods include AutoDAN~\cite{liu2024autodan}, which leverages genetic algorithms to produce fluent and stealthy adversarial suffixes, and its successor AutoDAN Turbo~\cite{liu2025autodanturbo}, which coordinates multiple LLMs for strategy development and attack evaluation. AdvPrompter~\cite{paulus2024advprompter} takes a different approach by fine-tuning a model specifically to generate coherent adversarial suffixes, enabling fast and automated jailbreaking.

On the defensive side, perplexity-based filtering~\cite{alon2023detecting} has proven effective in identifying adversarial suffixes by exploiting their typically high perplexity. However, newer attacks are designed to bypass such detection mechanisms by optimizing fluency and semantic plausibility.
In addition, work on language model inversion~\cite{morris2023languagemodelinversion} explores the recovery of original prompts from output probabilities, similar to reconstruction techniques in computer vision. These findings have informed strategies for generating adversarial prompts using only output distributions.

Unlike prior work focused on suffix generation or language inversion as an offensive tool, our research seeks to understand and mitigate these vulnerabilities.
In particular, we study \textbf{``evil twin'' prompts} as defined in~\citet{melamed2024prompts,rakotonirina2024evil}. Given a text prompt $\bx$ and the completion, $\by$, we performed an optimization so that given $\by$, we find a new nonsensical $\bxa$ -- the ``evil twin'' -- such that the loss $\Loss(\bxa,\by;\net) \ll \Loss(\bx,\by;\net)$, where $\Loss$ is the next-token prediction loss of the LLM and $\net$ are LLM’s parameters. These $\bxa$ are out-of-distribution, syntactically implausible inputs that nonetheless lead to the same output, as illustrated in \autoref{tab:evil_twins_examples}. Despite producing identical continuations, $\bx$ and $\bx^\star$ induce notably different
entropy distributions.
These prompts are also fragile -- small changes typically break the adversarial effect, highlighting a key vulnerability in LLM robustness and alignment.

\begin{table*}
\centering
\footnotesize
\resizebox{\textwidth}{!}{
\begin{tabular}{llc}\toprule
\textbf{Input} & \textbf{Output $\by$} & \textbf{Loss} \\
\midrule
%\cline{2-3}
$\bx$~: Stevens recorded and produced the album at multiple & \multirow{2}{*}{locations in the United}  & 5.3642 \\
$\bxa$: Zo Certified Cities (. broadcastquartered Fitness Academy thirteen   & &  \textbf{5.1302} \\
\midrule
$\bx$~: After the introduction of the Majors , The   & \multirow{2}{*}{British Army was divided}   & 11.2146 \\
$\bxa$: REQU Apart British received reformsMilitaryestic Division The  & &  \textbf{7.1899} \\
\midrule
$\bx$~: The founding director , Peggy Loar , left   & \multirow{2}{*}{the University of California}   & 7.2669\\
$\bxa$: tested UberERIC definitionCalifornia sustainability RutgersOL Jensen regarding  & &  \textbf{6.4402} \\
\midrule
$\bx$~: Ruiz notes that writing also has the power & \multirow{2}{*}{\centering to change the world} & 5.9135 \\
$\bxa$: Report Global feminism agenda Representatives tell Sacredixties Trying & & \textbf{4.6041} \\
\bottomrule
\end{tabular}
}
\caption{Original inputs $\bx$ and adversarial examples $\bxa$ generated using the GCG method for the SmolLM-360M model. The table shows how each original input and its corresponding adversarial example result in the same output, along with the loss values calculated for the output token IDs.
These examples show that LLMs can be manipulated into assigning lower loss to nonsensical prompts than to the original, meaningful input -- highlighting a vulnerability that ILM is designed to address.}
\label{tab:evil_twins_examples}
\end{table*}

To handle evil twin prompts, we propose Inverse Language Modeling (ILM), a novel training framework that improves LLM robustness. ILM enables \textbf{both forward modeling and partial inversion}, encouraging the model to not only generate fluent output but also remain sensitive to input semantics.

\section{Method}

\subsection{Preliminary Study on PAG on Text Classification}
In this preliminary experiment, we investigate the application of Perceptually Aligned Gradients (PAG)~\cite{ganz23aPAG} to sentence classification using hidden state representations from the DistilBERT language model~\cite{sanh2019distilbert}.  While PAG has been primarily explored in the context of image classification, we adapt the methodology to the hidden state space of a transformer model to explore its effects on robustness and interpretability in NLP tasks. The core idea of PAG is to encourage gradients to align with semantically meaningful directions, and we hypothesize that this can lead to more robust and interpretable text representations as well.

To prove our point, we ran a proof-of-concept experiment using a classifier trained on top of the hidden state associated with the \texttt{[CLS]} token,
adopting the \texttt{distilbert-base-multilingual-cased}~\cite{Sanh2019DistilBERTAD}, on Amazon Review Multi dataset~\cite{amazon-reviews-dataset}.
We considered 12 classes as the combination of some languages (English, German, Spanish, and French) and some review ratings (1, 3, and 5 stars).

\minisection{PAG Application}
To apply PAG, we modify the standard cross-entropy loss function with a regularization term that encourages alignment between the input gradient and a ``proxy'' ground-truth gradient. The modified loss function for the classifier built on top of the frozen DistilBERT backbone is:

\begin{equation}
\begin{split}
\Loss &= \Loss_{CE}(f_\theta(\bx), y) + \lambda\; \Loss_{PAG}(\bx)\quad\text{where}\\
\Loss_{PAG}(\bx) &= \frac{1}{C} \sum_{y = 1}^C 1 - \frac{\nabla_{\bh}f_{\theta}(\bx)_y^\top~g(\bx, y)}{\norm{\nabla_{\bh}f_{\theta}(\bx)_y} \norm{g(\bx, y)}}
\end{split}
\label{eq:pag_classification_loss}
\end{equation}
where:
\begin{itemize}
    \item \(\bx\) is the input sentence,
    \item \(y\) is the true class label,
    \item \(f_{\theta}(\bx)\) is the classifier that takes the DistilBERT hidden state (\(\bh\)) as input and predicts the class,
    \item \(\Loss_{CE}\) is the cross-entropy loss,
    \item \(\lambda\) is a hyperparameter controlling the strength of the PAG regularization,
    \item \(C\) is the number of classes,
    \item \(\nabla_{\bh}f_{\theta}(\bx)_y\) is the gradient of the classifier's output for class \(y\) with respect to the hidden state \(h\),
    \item \(g(\bx, y)\) is the ``proxy'' ground-truth gradient for \(y\).
\end{itemize}

\minisection{``Proxy'' Ground-Truth Gradient}
We define the ``proxy'' ground-truth gradient (\textbf{PAG} variant) as the difference between the hidden state of the input sentence $\bh_\bx$ and the hidden state of a randomly sampled sentence $\bu_y$ from the same class $y$:

\begin{equation}
g(\bx, y) = \bu_y - \bh_\bx.
\label{eq:pag}
\end{equation}

This encourages the model to learn hidden state representations where the gradient points in the direction of other examples from the same class.

Another variant for this $g(\cdot)$ function, named \textbf{Identity}, has been tested and compared. This one forced the model to reconstruct the input via the received gradients as:

\begin{equation}
g(\bx, y) = \bx.
\label{eq:pag-id}
\end{equation}

The baseline is the model trained with the same architecture and hyperparameters but $\lambda=0$, to exclude $\Loss_{PAG}$. 

\minisection{Evaluation}
According to the results in \autoref{tab:pag_variants_multiclass}, the strongest model in robustness is the one trained with the full PAG loss with \autoref{eq:pag}, which forces the model to make the gradients on the input point towards the direction of the predicted class. These models have been attacked by APGD, Square~\cite{croce2020reliable}, and FGSM~\cite{goodfellow2014explaining}.

\begin{table}[htbp]
\resizebox{\linewidth}{!}{
\begin{tabular}{lccccc}
\toprule
attack $\rightarrow$ &  \multicolumn{2}{c}{\textbf{APGD}} & \textbf{Square} & \multicolumn{2}{c}{\textbf{FGSM}} \\
&  \multicolumn{2}{c}{\citet{croce2020reliable}} & \citet{croce2020reliable} & \multicolumn{2}{c}{\citet{goodfellow2014explaining}} \\
& \multicolumn{2}{c}{$\varepsilon$} &  & \multicolumn{2}{c}{$\alpha$} \\
$g(\bx)$ $\downarrow$ & $1\text{e-}3$ & $0.5$ & &  $5\text{e-}3$ & $1\text{e-}2$ \\
\midrule
Baseline & 36.5\% & 31.2\% & 36.3\% &  27.3\% & 8.9\% \\
Identity & 28.3\% & 25.0\% & 27.2\% &  25.7\% & 8.0\% \\
\textbf{PAG} & \textbf{48.1\%} & \textbf{45.0\%} & 
\textbf{49.3\%} & \textbf{43.5\%} & \textbf{25.7\%} \\
\bottomrule
\end{tabular}
}
\caption{Robustness of classifier models with PAG variants under APGD, Square, and FGSM attacks. Higher percentages indicate stronger robustness.}
\label{tab:pag_variants_multiclass}
\end{table}

\subsection{Inverse Language Modeling}
Our procedure takes inspiration from robust classifiers that have Perceptually Aligned Gradients (PAG)~\cite{ganz23aPAG,mirza2024shedding} over the input space.
ILM is non-iterative and just requires double backpropagation~\cite{lecun92double}, which can be easily implemented with current autograd tools.

ILM performs the following: instead of training the LLM to \emph{only} maximize $p(\by|\bx)$,
we also invert it and from the output $\by$, we aim to reconstruct the input $\bx$.
This procedure is not simply a mere double forward pass with the original text and its reverse. Instead, we first impose a loss for $p(\by|\bx)$, yet instead of updating the weights, we also receive gradients over input tokens $\nabla_\bx \Loss(\bx,\by;\net)$ requiring them to predict some tokens in $\bx$, depending on the exact model variant among the ones discussed later.

This focus on bidirectional understanding during pretraining is key to improving the model's overall language comprehension. 

\begin{figure}
    \centering
    \includegraphics[width=\linewidth]{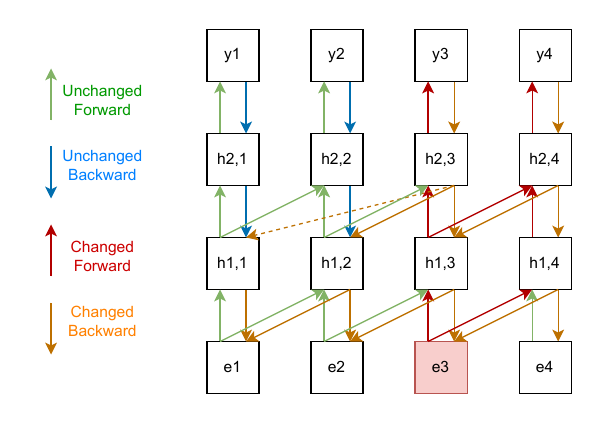}
    \caption{Gradient influence diagram in an LLM changing only token $e_3$. In the forward pass, only future hidden states are affected. In the backward pass, the change propagates to every embedding token.}
    \label{fig:llm_gradients_changing_embedding}
\end{figure}

From the gradients flow outlined in \autoref{fig:llm_gradients_changing_embedding},
it is possible to see that the influence of a single token affects only the future hidden states in the forward pass, while it influences every other token during the backward pass. This means that in the gradients received on the input tokens are affected by the entire sequence.

We will use ILM at training time, while at test time we exploit the GCG algorithm to find, given an original text prompt $\bx$ and the completion $\by$, a new nonsensical $\bxa$ such that the loss $\Loss(\bxa,\by;\net) \ll \Loss(\bx,\by;\net)$, where $\Loss$ is the next-token prediction loss of the LLM and $\net$ are LLM’s parameters. We prove there exists a nonsensical prompt $\bxa$ that ``connects'' better to $\by$ (lower loss) than the natural $\bx$.

The standard formulation of Perceptually Aligned Gradients (PAG), as typically applied in image classification -- \autoref{eq:pag_classification_loss} -- is not directly transferable to Large Language Models (LLMs). This is primarily due to the fundamental differences in the nature of the input data. Images are represented as tensors with continuous values, inherently containing class-discriminative information within a single sample. In contrast, sequence data processed by LLMs relies on the entire sequence context for prediction, not isolated elements. Furthermore, the input tokens are initially represented as one-hot vectors, which do not encode semantic information.

Calculating gradients with respect to these input tokens poses significant challenges:

\begin{itemize}[itemsep=2pt, leftmargin=*]
    \item Gradients with respect to individual input tokens lack the crucial context of the entire sentence, analogous to a single pixel being insufficient to determine an image's class.
    \item The vast vocabulary size of LLMs (hundreds of thousands of tokens) renders the class-iterative PAG loss computationally infeasible due to the sheer number of classes, exacerbating the entropy in the classification task.
\end{itemize}

Consequently, our approach deviates from the standard PAG for classifiers. We opt to focus solely on the gradients with respect to the actual input tokens and, furthermore, we intend to classify the actual tokens, as in \autoref{fig:inverse_lm_idea_schema}, rather than employing the cosine distance for gradient direction, aiming to leverage this for LLM inversion as well.

For this experiments, the architecture of our model is a small decoder-only transformer, with \textbf{Weight Tying}~\cite{press2017using,inan2017tying} enabled, with 3 hidden layers and an hidden layers vector size set to 640.

The dataset used is \textbf{TinyStories}~\cite{eldan2023tinystoriessmalllanguagemodels} with a tokenizer trained from scratch using the standard Byte-Pair Encoding~\cite{gage1994newbytepairencoding}, in order to have a flexible vocabulary size, to eventually have experiments of different complexities and entropy in the next token classification. Specifically, we used a vocabulary of 2048 possible tokens.
Also, the dataset samples include an overlap of 25\% between the original sentences. This overlap increases variability in sentence starts, providing the model with more diverse context patterns and better approximating realistic sentence-completion scenarios.
The finally constructed dataset
\footnote{\url{https://huggingface.co/datasets/DaveGabe/TinyStoriesV2_cleaned-voc2048-seq256-overlap25}}
has been uploaded to HuggingFace for reproducibility,
in pair with the associated tokenizer
\footnote{\url{https://huggingface.co/DaveGabe/TinyStoriesV2_cleaned-voc2048-seq256-overlap25-bpe-tokenizer}}.

The backward prediction strategy shares a common logic across all model variants, which can be formalized as follows. Given the cross-entropy loss between predicted tokens and their ground truth, we compute the gradient with respect to the embedding vectors. To handle different variants consistently, we define a mapping $\phi(\be_i, \nabla_{\be_i} \Loss_{CE})$ that specifies how the gradient is interpreted for classification.
Then, the output of $\phi(\dots)$ is normalized and used with the LM Head weight matrix to get a probability distribution over the vocabulary.
The general backward prediction for any token can then be written as:

\begin{equation}
\begin{split}
    \Loss_{CE} &= CE(\by_\text{true}, \by_\text{pred}) \\
    \bg_i &= \text{LayerNorm}(\phi(\be_i, \nabla_{\be_i} \Loss_{CE})) \\
    \bz_i &= \bW_\text{LM\_head} \; \bg_i \\
    \mathbf{\hat{y}}_i &= \text{softmax}(\bz_i)
\end{split}
\label{eq:inverse_lm_backward_prediction_general}
\end{equation}

\begin{figure}
    \centering
    \includegraphics[width=\linewidth]{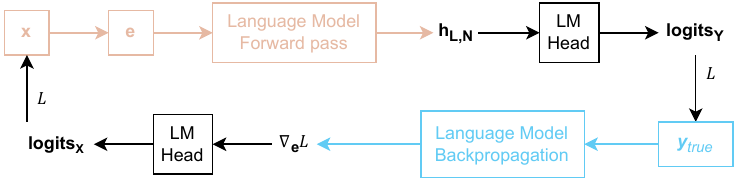}
    \caption{Parallelism between last hidden states and embedding gradients: both can be mapped through LM head to token predictions.}
    \label{fig:grad_lm_head_parallelism}
\end{figure}

This formulation highlights a very nice parallelism between the forward and backward mode, which is summarized in \autoref{fig:grad_lm_head_parallelism}:
the gradients vector on the embeddings of a specific token is used in the same way as the last hidden state that will conduct to a prediction of the next token.
Indeed, replacing $\nabla_{e_i}\Loss_{CE}$ with the last hidden state will give us exactly the standard forward pass of an LLM.
This is possible because the dimensionality of the last hidden state and the one of the embedding vector is the same.

The final loss used to train these models is obtained by the addition of the inverse LM prediction loss, where we observed $\lambda = 2.0$ to be a good hyperparameter:
\begin{equation}
\begin{split}
    \Loss =&
\underbrace{\Loss_{CE}(\by_\text{true}, \by_\text{pred})}_\text{Forward: from the input x, encode y} \\
+ &
\underbrace{\lambda\ \Loss_{CE}(\bx, \mathbf{\hat{y}})}_\text{Backward: from y, decode back x}
\end{split}
\end{equation}

\subsection{Model Variants}
We evaluate four training strategies, each differing in how gradients are used for inversion:

\textbf{Baseline.} The model is trained only with the standard forward Cross-Entropy loss.  

\textbf{Identity.} During training, the model is required to reconstruct every input token directly from its corresponding gradient:  
\begin{equation}
p(\bx_i \mid \nabla_{\be_i} \Loss_{\text{CE}}(f_\theta(\bx), \by)), \quad \forall i \in [1, N].
\label{eq:identity}
\end{equation}

\textbf{BERT-like.} A subset of tokens is masked in the input sequence, and the model must predict them from the embedding gradients -- analogous to the BERT~\cite{devlin-etal-2019-bert} training scheme, but applied in the backward pass.  

\textbf{Inv-First.} Only the first token of the sentence is reconstructed from its gradient, by predicting  
\begin{equation}
p(\bx_0 \mid \nabla_{\be_0} \Loss_{\text{CE}}(f_\theta(\text{\texttt{[PAD]}} \,\|\, \bx_{1:N}), \by)).
\label{eq:invfirst}
\end{equation}

Each strategy comes in two approaches, depending on the implementation of the $\phi(\dots)$ function mentioned in \autoref{eq:inverse_lm_backward_prediction_general}.
When we consider gradients as directions, we classify on $\be_i - \nabla_{\be_i} \loss_{CE}$, thus by imposing
$\phi(\be_i, \nabla_{\be_i} \Loss_{CE}) = \be_i - \nabla_{\be_i} \Loss_{CE}$.
On the other hand, the case where we use the gradients as pure values is simpler: $\phi(\be_i, \nabla_{\be_i} \Loss_{CE}) = \nabla_{\be_i} \Loss_{CE}$, discarding the input embedding value.
\section{Experimental Evaluation}
We evaluated the performance of these models from a twofold perspective:
their ability to invert a part of text, given the continuation but not what comes before; their robustness against GCG attack.
These aspects are equally important to develop robust and grounded LLMs.

\subsection{Inversion Procedure}
The evaluation relies on three complementary measures, all tailored to the inverse generation setting. Validation loss and validation accuracy are computed under the same conditions as training, serving as baseline indicators of model fit and predictive performance. In addition, we report Inverse LM accuracy, which measures the ability of a model to reconstruct a masked token $\bx_i$ from its gradients, given the remaining context. This provides a direct assessment of the backward prediction mechanism.

Formally, for an input sequence $\bx = (\bx_1, \ldots, \bx_N)$, the $i$-th token is replaced with a placeholder, yielding $\bx' = (\texttt{<|pad|>}, \bx_{i+1}, \ldots, \bx_N)$, with targets shifted as $\by' = (\bx_{i+1}, \ldots, \bx_N)$. The cross-entropy loss $\Loss_\text{CE}(\bx', \by'; \theta)$ produces gradients with respect to the masked embedding $\nabla_{\be_i}$, which are transformed via the backward prediction rule (\autoref{eq:inverse_lm_backward_prediction_general}) into a distribution $\hat{\by}_i$. Inverse LM accuracy is then computed by comparing $\argmax \hat{\by}_i$ against the ground-truth token $\bx_i$.

\subsection{Inversion Evaluation}
To assess inversion capabilities, we extend the task beyond single-token recovery and instead invert multiple tokens autoregressively.  
The procedure follows a beam-search strategy, as detailed in \autoref{alg:inversion_evaluation}, where candidate prefixes are iteratively expanded and filtered by perplexity until a coherent reconstruction emerges.
\begin{algorithm}[t]
    \caption{Autoregressive Inversion Evaluation with Beam Search}
    \label{alg:inversion_evaluation}
    \begin{algorithmic}[1]
        \Require Input sample $\bx$ of length $n$, beam size $b$, split position $k$
        \Ensure Inverted prefix $\bx_{\text{inv}}$
        \State $\bx_p \gets \bx_{0:k}$ \Comment{Original prefix (hidden)}
        \State $\bx_s \gets \bx_{k:n}$ \Comment{Visible suffix}
        \State $\bX \gets \{\bx_s\}$ \Comment{Initialize beam set with suffix only}
        \While{inverted prefix not sufficiently long}
            \For{each sequence $\bx' \in \bX$}
                \State Compute top-$b$ tokens for the previous position
                \State Extend $\bx'$ with each candidate token
            \EndFor
            \State $\bX \gets$ top-$b$ sequences with lowest perplexity
        \EndWhile
        \State \Return $\bx_{\text{inv}} \gets \arg\min_{\bx' \in \bX} \text{Perplexity}(\bx')$
    \end{algorithmic}
\end{algorithm}

In the evaluation process, we considered only the combination of initialization strategy and model variant used in the specific training process. This means that the \texttt{Identity} has the unknown token initialized using the simple bigram, while other variants have it set to \texttt{<|pad|>}, since they reflect the same initialization strategy used during training. Of course, we cannot initialize the \texttt{Identity} model during inversion with the real token, as in training, because we do not know it yet. However, the best approximation we can do is to use a bigram model, which is pretty simple but also powerful to help the model invert better than starting from a totally random token or using a fixed \texttt{<|pad|>} because it has never observed it during training.

In order to make this final evaluation on inversion as complete as possible, we introduced several new metrics: they allow us to better comprehend the obtained results and have a better understanding of the model training strategies applied. When we refer to a third-party model, we are using \texttt{meta-llama/Llama-3.2-1B}.
\footnote{\url{https://huggingface.co/meta-llama/Llama-3.2-1B}}
\begin{itemize}
    \item \emph{Rec} (\texttt{token\_recall}) refers to the fraction of unique tokens from the reference sequence that were correctly generated by the model. A higher recall value means the model captured more of the words from the reference
    \item \emph{Prec} (\texttt{token\_precision}) refers to the fraction of unique tokens in the generated sequence that are present in the reference. A higher precision value indicates the model didn't introduce many irrelevant or ``hallucinated'' words
    \item \emph{F1} (\texttt{token\_f1}) refers to the harmonic mean of precision and recall. It provides a single score that balances the trade-off between the two. A high F1 score indicates a good balance of both generating relevant tokens and avoiding irrelevant ones
    \item \emph{Acc} (\texttt{positional\_accuracy}) refers to the exact token match at each position in the generated sequence compared to the reference: unlike the token-based metrics above, this one is sensitive to token order
    \item \emph{OPP} (\texttt{original\_prefix\_perplexity}) measures the perplexity of the original prefix text alone, using the third-party model. This should serve as an indication of ``how natural'' the prefix text is. This metric will be \emph{the same} for all models, since it does not depend on the model, but only on the data to be predicted.
    \item \emph{FPP} (\texttt{full\_predicted\_perplexity}) measures the perplexity of the predicted prefix text, concatenated with the suffix, using the third-party model. This should serve as an indication of ``how grounded'' the generated prefix is with the suffix
    \item \emph{PPP} (\texttt{predicted\_prefix\_perplexity}) measures the perplexity of the predicted prefix text alone, using the third-party model. This should serve as an indication of ``how natural'' the generated text is
    \item \emph{SS} (\texttt{semantic\_similarity}) refers to the semantic meaning of the generated text compared to the ground-truth, regardless of the specific words used. It computes the cosine similarity between the embedding of the two sentences, obtained by an external model
        \footnote{\url{https://huggingface.co/sentence-transformers/all-MiniLM-L6-v2}}

\end{itemize}

\begin{table}[bthp]
\centering
\resizebox{\linewidth}{!}{
\begin{tabular}{rlcccc}
\toprule
           & \textbf{Grad.} & \textbf{Rec $\uparrow$}  & \textbf{Prec $\uparrow$} & \textbf{F1 $\uparrow$}   & \textbf{Acc $\uparrow$}  \\
\midrule
Baseline   &                & 20.9\%          & 18.8\%          & 19.7\%          & 2.4\%          \\
\midrule
Inv-First  &                & 11.3\%          & 10.1\%          & 10.7\%          & 1.7\%          \\
Bert-like  & Val.           & 2.9\%           & 2.7\%           & 2.8\%           & 0.3\%          \\
Identity   &                & 0.7\%           & 0.7\%           & 0.7\%           & 0.1\%          \\
\midrule
Inv-First  &                & 13.3\%          & 12.0\%          & 12.6\%          & 2.4\%          \\
Bert-like  & Dir.           & 0.1\%           & 0.1\%           & 0.1\%           & 0.1\%          \\
Identity   &                & \textbf{22.5\%} & \textbf{20.2\%} & \textbf{21.2\%} & \textbf{2.5\%} \\
\bottomrule
\end{tabular}
}
\caption{Inversion evaluation on token-level metrics (Recall, Precision, F1, Accuracy). Higher values mean better recovery of original tokens.}
\label{tab:ilm-evaluation-token-metrics}
\end{table}

\begin{table}[htbp]
\centering
\resizebox{\linewidth}{!}{
\begin{tabular}{rlccccc}
\toprule
           & \textbf{Grad.} & \textbf{OPP}  & \textbf{FPP $\downarrow$}   & \textbf{PPP $\downarrow$}  & \textbf{SS $\uparrow$}    \\
\midrule
Baseline   &                & 37.83           & \textbf{8.34}    & 112.82          & \underline{0.28} \\
\midrule
Inv-First  &                & 37.83           & 10.21            & 1576.23         & 0.25             \\
Bert-like  & Val.           & 37.83           & 11.54            & 5501.86         & 0.17             \\
Identity   &                & 37.83           & 13.88            & 14658.58        & 0.12             \\
\midrule
Inv-First  &                & 37.83           & 9.77             & 1012.80         & \textbf{0.30}    \\
Bert-like  & Dir.           & 37.83           & 11.05            & 563.26          & 0.11             \\
Identity   &                & 37.83           & \textbf{8.34}    & \textbf{106.31} & \textbf{0.30}    \\
\bottomrule
\end{tabular}
}
\caption{Sentence-level inversion metrics: OPP (original prefix perplexity), PPP (predicted prefix), FPP (full predicted), SS (semantic similarity).}
\label{tab:ilm-evaluation-sentences-metrics-llama}
\end{table}

Note that \emph{FPP} shows much less variance between the tested models, because it computes the perplexity of the entire sentence, which is the concatenation of the predicted prefix and the given suffix from the dataset. Since the latter is much longer than the former, the values of \emph{FPP} tend to be pretty low. However, we are interested in the difference between the models.

Also, it is clearly visible that \emph{PPP} is one order of magnitude larger in the models that predict the previous token using the gradient vector, without summing it up first with the embedding of the input token they're inverting on, used as the initialization value.
This clearly demonstrates the intuition for which using gradients as directions would have made the model hold much better.

In \autoref{appendix:inversion_qualitative_examples}, we provide some qualitative examples of inversion.

\subsection{Robustness Against GCG}
We evaluate robustness using the success rate of Greedy Coordinate Gradient (GCG) attacks~\cite{zou2023universal}.  
This benchmark aligns naturally with the objectives of our training procedure: our ultimate goal is to produce LLMs that are more grounded, ensuring that their responses are faithful to and informed by the prompts they receive.

The procedure to evaluate the models follows these rules, which are repeated for 30\% randomly picked samples from the test set, but consistently chosen for all the model variants:

\begin{algorithm}
    \caption{Single-Sentence GCG Attack}
    \label{alg:single_sentence_gcg}
    \begin{algorithmic}[1]
        \Require Expected continuation string $\by$ to be attacked,
                    length of the attack prefix $n$,
                    number of iterations $T$
        \Ensure Best attack prefix $\bxa$ with loss $\loss_\text{GCG}$
        
        \State $\bxa \gets$ random one-hot tokens matrix of size $|V| \times n$
        \State $step \gets 0$                       \Comment{Iteration counter}
        \State $d \gets 0$                          \Comment{Loss non-decrease counter}
        \State $\loss_{\text{old}} \gets \infty$    \Comment{Last loss found}
        
        \While{$step < T$ \textbf{and} $d < 10$}
            \State Compute a batch of candidate prefixes $\bX$ running one step of \textbf{GCG}
            \State $\bxa \gets \text{arg\;min}_{\bx \in \bX} \loss_\text{CE}(\bx, \by, \net)$
            \State $\loss_\text{GCG} \gets \loss_\text{CE}(\bxa, \by, \net)$    \Comment{Take the min loss so far}
            \If{$\loss_\text{GCG} < \loss_{\text{old}}$}
                \State $\loss_{\text{old}} \gets \loss_\text{GCG}$
                \State $d \gets 0$
            \Else
                \State $d \gets d + 1$
            \EndIf
            \State $step \gets step + 1$
        \EndWhile
        
        \State \Return $\loss_\text{GCG}$
    \end{algorithmic}
\end{algorithm}

\begin{table}[htbp]
\resizebox{\linewidth}{!}{
\begin{tabular}{rlcc}
\toprule
           & \multirow{2}{*}{\textbf{Grad.}} & \textbf{GCG}             & \textbf{GCG Average Steps}   \\
           &                                 & \textbf{Success Rate $\downarrow$} & \textbf{(mean ± stddev)}     \\
\midrule
Baseline   &                                 & 95.9\%                   & 277 $\pm$ 148                \\
\midrule
Identity   &                                 & 88.1\%                   & 274 $\pm$ 145                \\
Bert-like  & Val.                            & \textbf{0.8\%}           & 249 $\pm$ 148                \\
Inv-First  &                                 & 85.0\%                   & 320 $\pm$ 134                \\
\midrule
Identity   &                                 & \underline{82.8\%}       & 284 $\pm$ 141                \\
Bert-like  & Dir.                            & 85.5\%                   & 287 $\pm$ 143                \\
Inv-First  &                                 & 89.3\%                   & 313 $\pm$ 134                \\
\bottomrule
\end{tabular}
}
\caption{Success rates of GCG adversarial prompts against ILM variants. Lower values indicate stronger robustness.}
\label{tab:small_tinystories_gcg_results}
\end{table}

From the results observable in \autoref{tab:small_tinystories_gcg_results}, we can notice that the majority of the variants have an improvement in robustness against GCG attacks.
Looking at the specific variant that was flagged as the best one in the inversion task in the previous chapter, which is \emph{Identity (grad. direction)}, it allowed a reduction in the success rate of more than 13\%. 
This improvement can be attributed to the model's ability to better condition the continuation of a sentence with the actual prompt it was given as input (also referred to as ``\emph{grounded}''), which may lead to a more robust model.
However, there is the specific case of \emph{Bert-like (grad. value)}, which corresponds to the model variant that imitates BERT in the backward pass, masking some tokens and letting the model predict them directly, classifying on the received gradient on the PAD token.
This model scores an incredibly low GCG success rate, making us suppose that it may actually strongly go in the direction of adversarially robust models, at least on the gradient-based white-box GCG attack. Given the huge difference between the baseline and this variant, the experiments have been repeated from the initial training phase, but they gave us the same results as in the first run of the pipeline. For sure, this aspect will require a deeper investigation.

\begin{figure}[htbp]
    \centering
    \includegraphics[width=\linewidth]{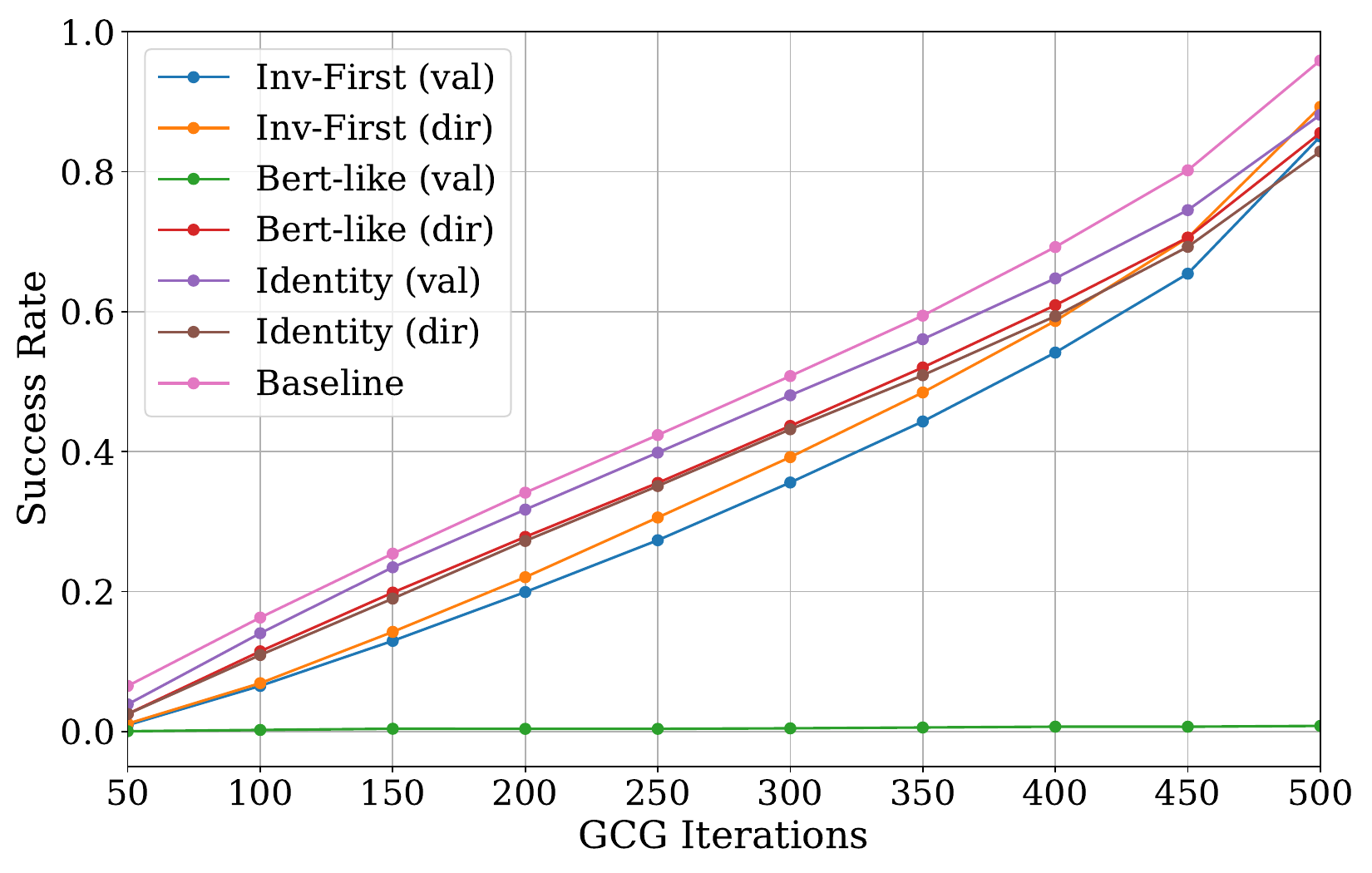}
    \caption{GCG Success Rate varying according to the number of iterations performed.}
    \label{fig:gcg_success_rate_varying_steps}
\end{figure}

In our experiments, we also studied the correlation between the number of GCG algorithm maximum allowed iterations and the success rate of the attack, always computed as the number of tokens that match between the LLM's responses to the original input $\bx$ and the attack input $\bx'$, while keeping all other hyperparameters, like the search window width, unaltered.
Interestingly, in \autoref{fig:gcg_success_rate_varying_steps} some lines actually cross each other when increasing the number of GCG iterations: this may indicate that some variants are more effective at different values of the GCG iterations. For instance, \emph{Inv-First (grad. direction)}, represented as the orange line, is better than \emph{Bert-like (grad. direction)}, represented as the red line, when the number of allowed iterations is pretty low; however, at the end of the plot, at the maximum number of iterations tested, their effectiveness is the opposite. That being said, this phenomenon is not dramatic and makes only slight changes in the final results reported in the previously discussed table.

To have a better understanding of the GCG results listed in the previous table,
we measured the following other metrics, always considering the subset of successful GCG attacks:
\begin{itemize}
    \item \emph{Original CE loss.} The cross-entropy loss on the unperturbed input,  
    $\Loss_{\text{CE}}(f_\theta(\bx_0, \ldots, \bx_N), \by_0, \ldots, \by_M)$
    where lower values indicate that $\by$ is a natural continuation of $\bx$.  

    \item \emph{Attack CE loss.} The cross-entropy loss on the adversarially perturbed input, $\Loss_{\text{CE}}(f_\theta(\bx'_0, \ldots, \bx'_N), \by_0, \ldots, \by_M)$
    where higher values are desirable, since a low loss would mean that $\by$ also appears natural given $\bx'$, which is precisely the vulnerability GCG exploits.  

    \item \emph{KL divergence.} The divergence between the output distributions of the original and perturbed inputs: $
    KL\big(f_\theta(\bx_0, \ldots, \bx_N), f_\theta(\bx'_0, \ldots, \bx'_N)\big)$.
    This measures how much the perturbation alters the probability distributions of the model’s predictions.
\end{itemize}
In all previous mathematical formulation, note that $f_\theta(\bx_0, ..., \bx_N)$ is the function which runs the LLM under attack and returns the logits that correspond to the prediction of the $\by$ output, not the ones that predict parts of the input prompt $\bx$ itself, as it would be wrong to consider for our analysis.

From the metrics in \autoref{tab:small_tinystories_ce_attack}, we observe that robust variants, such as \emph{Identity (grad. direction)}, not only exhibit a substantially lower attack success rate (ASR) compared to the baseline, but also display a smaller increase in Cross-Entropy loss when the attack succeeds.
Recall that this \emph{delta} quantifies the extent to which the model is ``fooled'' by the attack, defined as the difference between the loss on the original, human-readable input and the loss on the adversarially generated sequence.
A higher delta indicates greater susceptibility, as the model interprets the attack sequence as being more strongly aligned with the target continuation $\by$.
Finally, the \textbf{KL-divergence} allows us to observe how much the output distributions returned by the LLM differ between $\bx$ and $\bx'$. The more different they are, the better the model can discriminate between them, recognizing that they are actually two distinct and different pieces of input.

\begin{table}[htbp]
\centering
\resizebox{\linewidth}{!}{
\begin{tabular}{rlcccc}
\toprule
           & \multirow{2}{*}{\textbf{Grad.}} & \textbf{Original X} & \textbf{Attack X'} & \textbf{Delta}     & \textbf{KL}           \\
           &                                 & \textbf{CE-loss $\downarrow$}  & \textbf{CE-loss $\downarrow$}   & \textbf{CE-loss $\downarrow$} & \textbf{Divergence $\uparrow$} \\
\midrule
Baseline   &                                 & 13.28               & 10.97              & 2.31               & 2.19                  \\
\midrule
Identity   &                                 & 12.77               & 11.21              & 1.56               & 2.23                  \\
Bert-like  & Val.                            & 13.26               & 10.25              & 3.01               & \textbf{54.19}        \\
Inv-First  &                                 & \textbf{11.09}      & 9.72               & \underline{1.37}   & 2.44                  \\
\midrule
Identity   &                                 & 12.58               & 11.12              & 1.46               & 2.47                  \\
Bert-like  & Dir.                            & 11.49               & 10.34              & \textbf{1.15}      & 2.23                  \\
Inv-First  &                                 & \underline{11.21}   & 9.81               & 1.40               & 2.44                  \\
\bottomrule
\end{tabular}
}
\caption{Evaluation of the best attack input found using GCG.}
\label{tab:small_tinystories_ce_attack}
\end{table}

To have a complete evaluation, we adopt a similar approach to the one we used during inversion:
using a third-party model to compute some other statistics lets us abstract away from the biases in our LLMs.
Here, since the perplexity of the attack prefix is computed with a third-party independent model, it can easily return the real naturalness of the generated prefix, instead of being influenced by the attack itself and wrongly reporting that it will be even more natural than the human prefix.
Remember that we are considering only the successful attacks, ignoring the ones that have failed, since they are not useful to understand the quality of the attacks. Also, because of that, the results involving the \texttt{bert-like} variant using gradients as values will have a much smaller number of samples that participate in these metrics.

\begin{table}[htbp]
\centering
\resizebox{\linewidth}{!}{
\begin{tabular}{rlccc}
\toprule
           & \multirow{2}{*}{\textbf{Grad.}} & \textbf{Original X} & \textbf{Attack X'}    & \textbf{Semantic}     \\
           &                                 & \textbf{Perplexity} & \textbf{Perplexity $\downarrow$} & \textbf{Similarity $\uparrow$} \\
\midrule
Baseline   &                                 & 44.14               & 17344.04              & 0.13                  \\
\midrule
Identity   &                                 & 43.98               & \textbf{8322.25}      & \textbf{0.18}         \\
Bert-like  & Val.                            & 40.37               & 11817.21              & 0.11                  \\
Inv-First  &                                 & 44.81               & \underline{9431.09}   & \underline{0.16}      \\
\midrule
Identity   &                                 & 44.71               & 10929.21              & 0.15                  \\
Bert-like  & Dir.                            & 44.74               & 10611.09              & 0.13                  \\
Inv-First  &                                 & 43.50               & 12344.85              & 0.13                  \\
\bottomrule
\end{tabular}
}
\caption{Evaluation of the attack input prefix against the original input prefix for successful GCG attacks.}
\label{tab:small_tinystories_ce_attack_third_party}
\end{table}

Representative qualitative examples of these attacks are provided for reference in \autoref{appendix:gcg_qualitative_examples}.

\subsection{Forward Mode Evaluation}

\begin{table*}[htbp]
\resizebox{\linewidth}{!}{
\begin{NiceTabular}{rl|[tikz=dotted]X}
\toprule
           & \textbf{Grad.} & \textbf{Completion for ``One day,''} \\
\midrule
Baseline   &                & a little boy named Tim wanted to travel to a far mountain. He asked his dad for a raft, \\
\midrule
Identity   &                & a little girl named Lucy went to the park with her mom. Lucy liked to play on the swings \\
Bert-like  & Val.           & a little boy was walking in the park. He noticed a big, shiny object in the park. \\
Inv-First  &                & they pinch. They find gold. They take pictures of stars. \\
\midrule
Identity   &                & a little girl named Amy was playing outside. She saw a big tree and thought it was a toy. \\
Bert-like  & Dir.           & a little girl named Lucy was playing in the garden. She saw a shiny ring on a branch. \\
Inv-First  &                & hey went to the beach with his mom. He saw something shiny and strange inside. \\
\bottomrule
\end{NiceTabular}
}
\caption{Example completion for the given prompt, in forward mode.}
\label{tab:inverse_lm__forward_mode_execution}
\end{table*}

\begin{table}[htbp]
\footnotesize
\centering
\begin{tabular}{rlcc}
\toprule
           & \textbf{Grad.} & \textbf{Perplexity $\downarrow$} & \textbf{CE Loss $\downarrow$} \\
\midrule
Baseline   &                & \textbf{4.83}         & \textbf{1.58}      \\
\midrule
Identity   &                & \underline{5.07}      & \underline{1.63}   \\
Bert-like  & Val.           & 5.79                  & 1.76               \\
Inv-First  &                & 8.41                  & 2.13               \\
\midrule
Identity   &                & \underline{5.08}      & \underline{1.62}   \\
Bert-like  & Dir.           & 5.42                  & 1.69               \\
Inv-First  &                & 6.82                  & 1.92               \\
\bottomrule
\end{tabular}
\caption{Quantitative forward mode evaluation.}
\label{tab:inverse_lm__forward_mode_quantitative}
\end{table}

Finally, we tested that these models are still functioning properly in forward mode, without experiencing \textbf{performance degradation}. Checking out the perplexity during training and validation, it has been observed that the performance of the custom models with the regularization term on the gradients $\nabla_\be\loss_\text{CE}$ does not penalize the model's ability to speak fluently during the standard usage in forward mode. 
This outcome is noteworthy, as adversarial training in literature often necessitates additional parameters or extended training to achieve comparable forward-mode performance, since part of the model’s capacity is effectively devoted to satisfying the adversarial objective.
In \autoref{tab:inverse_lm__forward_mode_quantitative} we can observe that the worst model is \emph{Inv-First (grad. value)}.
This relatively high perplexity value is also confirmed in the qualitative examples in the tables below, where the sentences are by far the ones that make less sense and seem more confused and repetitive.

\section{Conclusions and Future Work}
In conclusion, this paper introduces Inverse Language Modeling (ILM) as a novel framework designed to simultaneously address two critical challenges in Large Language Models: robustness and grounding.
Our experiments demonstrate ILM's potential to enhance LLMs' resilience against input perturbations, a key step towards mitigating vulnerabilities to adversarial attacks.
Furthermore, ILM offers a pathway to improved grounding, enabling LLMs to better correlate their outputs with the input prompts and thereby facilitating the identification of potentially problematic input triggers.
With LLMs transforming into agents that interpret user input and take real actions, ILMs may offer a pathway toward new methods for building language models that are fundamentally more reliable, controllable, and trustworthy.

\minisection{Future work}
There are several promising avenues for future research.
While ILM is introduced within the context of pre-training, an interesting direction would be to explore its application in the \textbf{fine-tuning stage}.  Specifically, one could investigate how the principles of inverse modeling can be incorporated into the fine-tuning process to improve the robustness and generalization of LLMs on downstream tasks.  Additionally, research could explore the potential benefits of combining ILM with instruction tuning, to further align LLM behavior with human preferences and instructions.
Future work should evaluate ILM on larger-scale LLMs, including Llama-3.2-1B, 7B, and even bigger models, to rigorously assess its scalability and effectiveness as model capacity increases.

\section*{Acknowledgements}
This work was supported by projects PNRR MUR PE0000013FAIR under the MUR National Recovery and Resilience Plan funded by the European
Union - NextGenerationEU and PRIN 2022 project 20227YET9B “AdVVent” CUP
code B53D23012830006. It was also partially supported by Sapienza research projects
“Prebunking”, “Adagio”, and “Risk and Resilience factors in disadvantaged young people: a multi-method study in ecological and virtual environments”. Computing was supported by CINECA cluster under projects Ge-Di HP10CRPUVC, EHPC-DEV-2025D06-096 and the Sapienza Computer Science Department cluster.
\section*{Ethics and Impact Statement}
The advancement of LLMs carries potential ethical implications,
especially in the era of agents.
Our investigation into this phenomenon contributes to a better understanding of how LLMs work and, thus, ultimately, to make them safer and more predictable. We believe that the publication of our research will promote a broader discussion on the responsible development of LLMs and contribute to the development of better defense mechanisms, as similar progress has already been made in the field of deep classifiers.

\bibliography{bibliography}

\begin{thebibliography}{28}
\providecommand{\natexlab}[1]{#1}
\providecommand{\url}[1]{\texttt{#1}}
\expandafter\ifx\csname urlstyle\endcsname\relax
  \providecommand{\doi}[1]{doi: #1}\else
  \providecommand{\doi}{doi: \begingroup \urlstyle{rm}\Url}\fi

\bibitem[Alon \& Kamfonas(2023)Alon and Kamfonas]{alon2023detecting}
Alon, G. and Kamfonas, M.
\newblock Detecting language model attacks with perplexity, 2023.
\newblock URL \url{https://arxiv.org/abs/2308.14132}.

\bibitem[Bender et~al.(2021)Bender, Gebru, McMillan-Major, and Shmitchell]{bender2021dangers}
Bender, E.~M., Gebru, T., McMillan-Major, A., and Shmitchell, S.
\newblock On the dangers of stochastic parrots: Can language models be too big?
\newblock In \emph{ACM conference on fairness, accountability, and transparency}, 2021.

\bibitem[Carlini et~al.(2024)Carlini, Jagielski, Choquette-Choo, Paleka, Pearce, Anderson, Terzis, Thomas, and Tram{\`e}r]{carlini2024poisoning}
Carlini, N., Jagielski, M., Choquette-Choo, C.~A., Paleka, D., Pearce, W., Anderson, H., Terzis, A., Thomas, K., and Tram{\`e}r, F.
\newblock Poisoning web-scale training datasets is practical.
\newblock In \emph{IEEE Symposium on Security and Privacy (SP)}, 2024.

\bibitem[Croce \& Hein(2020)Croce and Hein]{croce2020reliable}
Croce, F. and Hein, M.
\newblock Reliable evaluation of adversarial robustness with an ensemble of diverse parameter-free attacks.
\newblock In \emph{ICML}, 2020.

\bibitem[Devlin et~al.(2019)Devlin, Chang, Lee, and Toutanova]{devlin-etal-2019-bert}
Devlin, J., Chang, M.-W., Lee, K., and Toutanova, K.
\newblock {BERT}: Pre-training of deep bidirectional transformers for language understanding.
\newblock In Burstein, J., Doran, C., and Solorio, T. (eds.), \emph{Proceedings of the 2019 Conference of the North {A}merican Chapter of the Association for Computational Linguistics: Human Language Technologies, Volume 1 (Long and Short Papers)}, pp.\  4171--4186, Minneapolis, Minnesota, June 2019. Association for Computational Linguistics.
\newblock \doi{10.18653/v1/N19-1423}.
\newblock URL \url{https://aclanthology.org/N19-1423/}.

\bibitem[Drucker \& Le~Cun(1992)Drucker and Le~Cun]{lecun92double}
Drucker, H. and Le~Cun, Y.
\newblock Improving generalization performance using double backpropagation.
\newblock \emph{IEEE Transactions on Neural Networks}, 3\penalty0 (6):\penalty0 991--997, 1992.
\newblock \doi{10.1109/72.165600}.

\bibitem[Ebrahimi et~al.(2018)Ebrahimi, Rao, Lowd, and Dou]{ebrahimi2018hotflip}
Ebrahimi, J., Rao, A., Lowd, D., and Dou, D.
\newblock Hotflip: White-box adversarial examples for text classification, 2018.
\newblock URL \url{https://arxiv.org/abs/1712.06751}.

\bibitem[Eldan \& Li(2023)Eldan and Li]{eldan2023tinystoriessmalllanguagemodels}
Eldan, R. and Li, Y.
\newblock Tinystories: How small can language models be and still speak coherent english?, 2023.
\newblock URL \url{https://arxiv.org/abs/2305.07759}.

\bibitem[Gage(1994)]{gage1994newbytepairencoding}
Gage, P.
\newblock A new algorithm for data compression.
\newblock \emph{The C Users Journal}, 12\penalty0 (2):\penalty0 23--38, 1994.

\bibitem[Ganz et~al.(2023)Ganz, Kawar, and Elad]{ganz23aPAG}
Ganz, R., Kawar, B., and Elad, M.
\newblock Do perceptually aligned gradients imply robustness?
\newblock In \emph{ICML}, 2023.

\bibitem[Goodfellow et~al.(2015)Goodfellow, Shlens, and Szegedy]{goodfellow2014explaining}
Goodfellow, I., Shlens, J., and Szegedy, C.
\newblock Explaining and harnessing adversarial examples.
\newblock In \emph{ICLR}, 2015.

\bibitem[Inan et~al.(2017)Inan, Khosravi, and Socher]{inan2017tying}
Inan, H., Khosravi, K., and Socher, R.
\newblock Tying word vectors and word classifiers: A loss framework for language modeling.
\newblock In \emph{ICLR}, 2017.

\bibitem[Keung et~al.(2020)Keung, Lu, Szarvas, and Smith]{amazon-reviews-dataset}
Keung, P., Lu, Y., Szarvas, G., and Smith, N.~A.
\newblock The multilingual amazon reviews corpus.
\newblock In \emph{EMNLP}, 2020.

\bibitem[Li et~al.(2024)Li, Li, Li, Lee, Cui, and Hu]{li2024faster}
Li, X., Li, Z., Li, Q., Lee, B., Cui, J., and Hu, X.
\newblock Faster-gcg: Efficient discrete optimization jailbreak attacks against aligned large language models, 2024.
\newblock URL \url{https://arxiv.org/abs/2410.15362}.

\bibitem[Liu et~al.(2024)Liu, Xu, Chen, and Xiao]{liu2024autodan}
Liu, X., Xu, N., Chen, M., and Xiao, C.
\newblock Autodan: Generating stealthy jailbreak prompts on aligned large language models, 2024.
\newblock URL \url{https://arxiv.org/abs/2310.04451}.

\bibitem[Liu et~al.(2025)Liu, Li, Suh, Vorobeychik, Mao, Jha, McDaniel, Sun, Li, and Xiao]{liu2025autodanturbo}
Liu, X., Li, P., Suh, G.~E., Vorobeychik, Y., Mao, Z., Jha, S., McDaniel, P., Sun, H., Li, B., and Xiao, C.
\newblock Auto{DAN}-turbo: A lifelong agent for strategy self-exploration to jailbreak {LLM}s.
\newblock In \emph{The Thirteenth International Conference on Learning Representations}, 2025.
\newblock URL \url{https://openreview.net/forum?id=bhK7U37VW8}.

\bibitem[Melamed et~al.(2024)Melamed, McCabe, Wakhare, Kim, Huang, and Boix-Adser{\`a}]{melamed2024prompts}
Melamed, R., McCabe, L., Wakhare, T., Kim, Y., Huang, H.~H., and Boix-Adser{\`a}, E.
\newblock Prompts have evil twins.
\newblock In \emph{EMNLP}, 2024.

\bibitem[Mirza et~al.(2024)Mirza, Briglia, Beadini, and Masi]{mirza2024shedding}
Mirza, M.~H., Briglia, M.~R., Beadini, S., and Masi, I.
\newblock Shedding more light on robust classifiers under the lens of energy-based models.
\newblock In \emph{ECCV}, 2024.

\bibitem[Morris et~al.(2023)Morris, Zhao, Chiu, Shmatikov, and Rush]{morris2023languagemodelinversion}
Morris, J.~X., Zhao, W., Chiu, J.~T., Shmatikov, V., and Rush, A.~M.
\newblock Language model inversion, 2023.
\newblock URL \url{https://arxiv.org/abs/2311.13647}.

\bibitem[Paulus et~al.(2024)Paulus, Zharmagambetov, Guo, Amos, and Tian]{paulus2024advprompter}
Paulus, A., Zharmagambetov, A., Guo, C., Amos, B., and Tian, Y.
\newblock Advprompter: Fast adaptive adversarial prompting for llms, 2024.
\newblock URL \url{https://arxiv.org/abs/2404.16873}.

\bibitem[Press \& Wolf(2017)Press and Wolf]{press2017using}
Press, O. and Wolf, L.
\newblock Using the output embedding to improve language models.
\newblock In \emph{ACL}, 2017.

\bibitem[Rakotonirina et~al.(2024)Rakotonirina, Kervadec, Franzon, and Baroni]{rakotonirina2024evil}
Rakotonirina, N.~C., Kervadec, C., Franzon, F., and Baroni, M.
\newblock Evil twins are not that evil: Qualitative insights into machine-generated prompts.
\newblock \emph{arXiv preprint arXiv:2412.08127}, 2024.

\bibitem[Sanh et~al.(2019{\natexlab{a}})Sanh, Debut, Chaumond, and Wolf]{Sanh2019DistilBERTAD}
Sanh, V., Debut, L., Chaumond, J., and Wolf, T.
\newblock Distilbert, a distilled version of bert: smaller, faster, cheaper and lighter.
\newblock \emph{ArXiv}, abs/1910.01108, 2019{\natexlab{a}}.

\bibitem[Sanh et~al.(2019{\natexlab{b}})Sanh, Debut, Chaumond, and Wolf]{sanh2019distilbert}
Sanh, V., Debut, L., Chaumond, J., and Wolf, T.
\newblock Distilbert, a distilled version of bert: smaller, faster, cheaper and lighter.
\newblock In \emph{NeurIPS}, 2019{\natexlab{b}}.

\bibitem[Vaswani et~al.(2017)Vaswani, Shazeer, Parmar, Uszkoreit, Jones, Gomez, Kaiser, and Polosukhin]{vaswani2017attention}
Vaswani, A., Shazeer, N., Parmar, N., Uszkoreit, J., Jones, L., Gomez, A.~N., Kaiser, {\L}., and Polosukhin, I.
\newblock Attention is all you need.
\newblock \emph{NeurIPS}, 30, 2017.

\bibitem[Xhonneux et~al.(2024)Xhonneux, Sordoni, G{\"u}nnemann, Gidel, and Schwinn]{xhonneux2024efficient}
Xhonneux, S., Sordoni, A., G{\"u}nnemann, S., Gidel, G., and Schwinn, L.
\newblock Efficient adversarial training in llms with continuous attacks.
\newblock In \emph{NeurIPS}, 2024.

\bibitem[Zhao et~al.(2024)Zhao, Zheng, Cai, Do, Kawaguchi, Goyal, and Shieh]{zhao2024accelerating}
Zhao, Y., Zheng, W., Cai, T., Do, X.~L., Kawaguchi, K., Goyal, A., and Shieh, M.
\newblock Accelerating greedy coordinate gradient and general prompt optimization via probe sampling, 2024.
\newblock URL \url{https://arxiv.org/abs/2403.01251}.

\bibitem[Zou et~al.(2023)Zou, Wang, Carlini, Nasr, Kolter, and Fredrikson]{zou2023universal}
Zou, A., Wang, Z., Carlini, N., Nasr, M., Kolter, J.~Z., and Fredrikson, M.
\newblock Universal and transferable adversarial attacks on aligned language models.
\newblock \emph{arXiv preprint arXiv:2307.15043}, 2023.

\end{thebibliography}
\bibliographystyle{icml2025}

\appendix
\onecolumn
\clearpage
\section{Inversion Qualitative Examples}
\label{appendix:inversion_qualitative_examples}

\renewcommand{\arraystretch}{1.3}

In the following tables, we can observe the difference in results between the baseline, the best variant (Identity, using gradients as directions), and the same variant but using the gradients as values, in addition to all other tested variants.

\begin{table}[htbp]
    \centering
    \footnotesize

    \begin{subtable}{\linewidth}
        \centering
        \begin{NiceTabular}{ll|[tikz=dotted]X}
            \toprule
            $\bx$  &  &  dad in the garden. He gives her a small shovel and a bag of bulbs. \\
            \midrule
            $\bxa$ Baseline & &  to play with his cars, and look at the shake. She feels on her hand. \\
            \midrule
            $\bxa$ Inv-First & (Val.) & zzle spowerlizza in her plate. She start to fence and leaves. \\
            \hdashline
            $\bxa$ Bert-like & (Val.) &  could buildDven measure its neighbign, how he sees nostiff. \\
            \hdashline
            $\bxa$ Identity & (Val.)  & Kugct propide,RallashQilndmawkeycessUuhingask do. \\
            \midrule
            $\bxa$ Inv-First & (Dir.) &  too hurt the car's bricket. It did not want to grow in a cage. \\
            \hdashline
            $\bxa$ Bert-like & (Dir.) &  Tim! Tim,ide, Sue, Sue, Tim!ide, "Tim, "Tim,ice. Tim! Tim!ittenbbed Tim! Tim,ide,auseectle. \\
            \hdashline
            $\bxa$ Identity & (Dir.)  &  cars, and gets on his hand. But he does not want to play with the towers. \\
            \midrule
            $\by$  &   & Bulbs are like round seeds that grow into flowers. Lily digs holes in the dirt and puts the bulbs inside. She covers them with more [...] \\
            \bottomrule
        \end{NiceTabular}
        \vspace{0.15cm}
        \caption{Inversion example of sample no. 1}
    \end{subtable}
    \vspace{0.25cm}
\end{table}

\begin{table}[htbp]
    \centering % Center the entire set of subtables
    \footnotesize
    \ContinuedFloat
    \begin{subtable}{\textwidth}
        \centering
        \begin{NiceTabular}{ll|[tikz=dotted]X}
            \toprule
            $\bx$   & &  play in the sand. They had a big bucket and a small shovel. They wanted to \\
            \midrule
            $\bxa$ Baseline & &  on his arm. He were playing with their cars, and looked at the window. They wanted to \\
            \midrule
            $\bxa$ Inv-First & (Val.) &  ovraph spower. Ben nodded. It looked happy and crunty, trying to \\
            \hdashline
            $\bxa$ Bert-like & (Val.) & stfister pink fire Fvery build loud budd poster watched closer before he heard someone \\
            \hdashline
            $\bxa$ Identity & (Val.)  &  canraask sn our saidJribistanezzemex-andight.ke work not \\
            \midrule
            $\bxa$ Inv-First & (Dir.) & rent. He walked closards the door. Lily and Ben were tragivous. They wanted to \\
            \hdashline
            $\bxa$ Bert-like & (Dir.) & ide, Sue,ittenect Tim!ide, Tim!ide,ide,ide,ittenice. Tim! Tim! Tim! Tim,ectbbedauseide, \\
            \hdashline
            $\bxa$ Identity & (Dir.)  &  her cars, and gets on the window. It was playing with their blocks. They wanted to \\
            \midrule
            $\by$  & & make a castle. They dug and piled and shaped the sand. They found some shells and stones to decorate their castle. "Look, our castle is [...] \\
            \bottomrule
        \end{NiceTabular}
        \vspace{0.15cm}
        \caption{Inversion example of sample no. 2}
    \end{subtable}
    \vspace{0.25cm}
\end{table}

\begin{table}[htbp]
    \centering
    \footnotesize
    \ContinuedFloat
    \begin{subtable}{\textwidth}
        \centering
        \begin{NiceTabular}{ll|[tikz=dotted]X}
            \toprule
            $\bx$   & &  They like to play in the park. They see a big swing. Lily wants to swing on it. She \\
            \midrule
            $\bxa$ Baseline  & &  a shake. It feels on his arm. He wants to play in the window. She \\
            \midrule
            $\bxa$ Inv-First & (Val.) & hed at being a good brush. Amy felt sorry for herself. she swings threve. She \\
            \hdashline
            $\bxa$ Bert-like & (Val.) &  poster how good. Ostfast tea, ride, seen swow, three, fide bit \\
            \hdashline
            $\bxa$ Identity & (Val.)  &  friend, sorished "Ochirt oversed "ownftittuggestign gulim way str \\
            \midrule
            $\bxa$ Inv-First & (Dir.) &  Tom won't wear a big small blue side. Lila does not want to go too far grass. She \\
            \hdashline
            $\bxa$ Bert-like & (Dir.) &  Sue, Sue, Sue,ittenittenittenice.ide, Tim! Tim!itten Tim! Tim! Tim,bbedause Tim!ide,ectle. \\
            \hdashline
            $\bxa$ Identity & (Dir.)  & er, but he does not want to play in the window. It feels on her arm. She \\
            \midrule
            $\by$   &  & runs to the swing and sits on it. "Push me, Ben!" Lily says. "Push me high!" Ben pushes Lily on the swing. Lily feels happy. [...] \\
            \bottomrule
        \end{NiceTabular}
        \vspace{0.15cm}
        \caption{Inversion example of sample no. 3}
    \end{subtable}
    \vspace{0.25cm}
\end{table}

\begin{table}[htbp]
    \centering
    \footnotesize
    \ContinuedFloat
    \begin{subtable}{\textwidth}
        \centering
        \begin{NiceTabular}{ll|[tikz=dotted]X}
            \toprule
            $\bx$   & &  with their toy cars in the living room. They liked to make noises and pretend they were driving \\
            \midrule
            $\bxa$ Baseline & &  his cars, and looked at the shake. They were playing with her hand. It was too \\
            \midrule
            $\bxa$ Inv-First & (Val.) &  laat. He ran towards the tight. Ben saw a big car crack. Tom was strong and \\
            \hdashline
            $\bxa$ Bert-like & (Val.) &  sailllaster more scatterfren, how could expl his dad drove himself too \\
            \hdashline
            $\bxa$ Identity & (Val.)  & oundddum cat.sideex picturesighU or promised s angry. "That's hearAH onore mommy \\
            \midrule
            $\bxa$ Inv-First & (Dir.) &  makes a big mess!" Bax did not like my toy car. Their cars started can make go fast \\
            \hdashline
            $\bxa$ Bert-like & (Dir.) &  "Tim, "Tim, Tim! Sue, Sue, Sue, Sue, Sue, Sue,ittenectitten Tim! Tim!ide, Tim,auseectbbedet, \\
            \hdashline
            $\bxa$ Identity & (Dir.)  &  he does not want to play in the window. He became playing with their blocks. They were very \\
            \midrule
            $\by$   & & fast. Lily had a pink car and Tom had a blue car. "Look, my car is faster than yours!" Tom said, zooming past Lily. "No, [...] \\
            \bottomrule
        \end{NiceTabular}
        \vspace{0.15cm}
        \caption{Inversion example of sample no. 4}
    \end{subtable}
    \vspace{0.25cm}

    \caption{Qualitative samples across all the model variants}
\end{table}

\clearpage

\renewcommand\tabularxcolumn[1]{m{#1}}

\section{GCG Attacks Qualitative Examples}
\label{appendix:gcg_qualitative_examples}

\renewcommand{\arraystretch}{1.3}

In the following tables, we can observe some examples of the attack strings found by applying the GCG algorithm on the tested model variants.
Although they do not show attack strings that carry out the meaning of a $\by$ continuation, and they are still meaningless,
our training procedure allowed us to have a more robust model, since the attack success rate is lower in all the variants; some having a better result than others, but all of them in some way beat the baseline.

\begin{table}[htbp]
\centering
\footnotesize
\begin{tabularx}{\linewidth}{lXXc}\toprule
 & \textbf{Input} & \textbf{Output $\by$} & \textbf{Loss} \\
\midrule
$\bx$~: & cry. He wished he had listened to his mom. He stared at the earth and hoped someone would & \multirow{3}{*}{ help him. Suddenly, he heard a voice }  & 11.55 \\
$\bxa$: & Boboach att ag ChS asked cake.vel near pftautautue. thoughtautaut smart would & &  \textbf{ 10.11 } \\
\midrule
$\bx$~: & long time, ever since the house was abandoned. I have many stories to tell & \multirow{3}{*}{ you. Would you like to hear them? }  & 17.22 \\
$\bxa$: & gga,Dveryng Do magic belie the theationationicy black name never sayirt<unk> & &  \textbf{ 12.99 } \\
\bottomrule
\end{tabularx}
\caption{Original inputs $\bx$ and adversarial examples $\bxa$ generated attacking with GCG the \texttt{Baseline} model.}
\vspace{0.5cm}
\end{table}

\begin{table}[htbp]
\centering
\footnotesize
\begin{tabularx}{\linewidth}{lXXc}\toprule
 & \textbf{Input} & \textbf{Output $\by$} & \textbf{Loss} \\
\midrule
$\bx$~: & Lily and Ben were friends who liked to play outside. But they did not like the same things. Lily & \multirow{3}{*}{ liked to make snowmen and snow angel }  & 13.22 \\
$\bxa$: & Lucy. Speez herself angO piecle you."lly named nexird opened cake".o.ter carrotmy & &  \textbf{ 12.14 } \\
\midrule
$\bx$~: & the ducks and the swings," Sam said."OK, Sam. That sounds fun," & \multirow{2}{*}{ Lily said.They opened the door and ste }  & 17.06 \\
$\bxa$: & Mommy than They TIon. up.Cleaseily am castle omisent. asked. crabo." & &  \textbf{ 13.85 } \\
\bottomrule
\end{tabularx}
\caption{Original inputs $\bx$ and adversarial examples $\bxa$ generated attacking with GCG the \texttt{Identity (grad. value)} model.}
\vspace{0.5cm}
\end{table}

\begin{table}[htbp]
\centering
\footnotesize
\begin{tabularx}{\linewidth}{lXXc}\toprule
 & \textbf{Input} & \textbf{Output $\by$} & \textbf{Loss} \\
\midrule
$\bx$~: & Lily and Ben were twins who liked to go on tours with their mom and dad. They & \multirow{3}{*}{ saw many new things and learned a lot. But }  & 24.14 \\
$\bxa$: & K forgot Tomriend rot became want or<unk><unk><unk> Onceriend sky. butterXgetherby & &  \textbf{ 23.08 } \\
\midrule
$\bx$~: & read books or draw pictures. They wanted to go outside and have fun. They looked at each other and & \multirow{3}{*}{ had an idea."Mom, we need to }  & 18.80 \\
$\bxa$: & Lis conizz Maskock!" t angry had nois What oyM sleep Benirdixfe & &  \textbf{ 18.27 } \\
\bottomrule
\end{tabularx}
\caption{Original inputs $\bx$ and adversarial examples $\bxa$ generated attacking with GCG the \texttt{Bert-like (grad. value)} model.}
\end{table}

\begin{table}[htbp]
\centering
\footnotesize
\begin{tabularx}{\linewidth}{lXXc}\toprule
 & \textbf{Input} & \textbf{Output $\by$} & \textbf{Loss} \\
\midrule
$\bx$~: & Lila and Ben were adventurous kids. They liked to explore new places and try new things. & \multirow{3}{*}{ One day, they went to the beach with their mom }  & 18.64 \\
$\bxa$: & so rest come so an One rock o per who<s> fr balling, M cars Whisk listen. likes Spot. & &  \textbf{ 13.47 } \\
\midrule
$\bx$~: & One day, a little boy named Tim found a peanut on the ground. He picked it up and & \multirow{3}{*}{ showed it to his mom. "Mom, what is this?" }  & 19.36 \\
$\bxa$: & pudd likesnn.hing It Sara "It where M whereanc It where pen where some?" madei & &  \textbf{ 17.06 } \\
\bottomrule
\end{tabularx}
\caption{Original inputs $\bx$ and adversarial examples $\bxa$ generated attacking with GCG the \texttt{Inv-First (grad. value)} model.}
\vspace{0.5cm}
\end{table}

\begin{table}[htbp]
\centering
\footnotesize
\begin{tabularx}{\linewidth}{lXXc}\toprule
 & \textbf{Input} & \textbf{Output $\by$} & \textbf{Loss} \\
\midrule
$\bx$~: & ." She is happy. She made a new pattern.She moves on to the flowers. & \multirow{3}{*}{ She says, "One, two, three, f }  & 11.15 \\
$\bxa$: & e Sheleix ey,se, amix4<unk>t. says, Herix help, in pawixt. & &  \textbf{ 9.55 } \\
\midrule
$\bx$~: & Imagine standing in a park with many trees and a big green lawn. & \multirow{3}{*}{ In the middle of the park is }  & 16.43 \\
$\bxa$: & fere re'age  att ApenXapard. thankarn vo park.XX keyft park. & &  \textbf{ 11.73 } \\
\bottomrule
\end{tabularx}
\caption{Original inputs $\bx$ and adversarial examples $\bxa$ generated attacking with GCG the \texttt{Identity (grad. direction)} model.}
\vspace{0.5cm}
\end{table}

\begin{table}[htbp]
\centering
\footnotesize
\begin{tabularx}{\linewidth}{lXXc}\toprule
 & \textbf{Input} & \textbf{Output $\by$} & \textbf{Loss} \\
\midrule
$\bx$~: & of plastic and has perfume on it. It does not grow or need water. It is & \multirow{3}{*}{ not real."Lily was confused and }  & 18.72 \\
$\bxa$: & everything her small, from pus Tim ThatF<unk> LV paper I h listen LI with. do & &  \textbf{ 15.15 } \\
\midrule
$\bx$~: & reat-grandpa. He was a brave explorer. He wore me on his & \multirow{3}{*}{ adventures."Lily was amazed. She wanted to }  & 20.26 \\
$\bxa$: & by ta anotherizCcessious rcessly, Lious tower her ask derirthdayI its & &  \textbf{ 17.54 } \\
\bottomrule
\end{tabularx}
\caption{Original inputs $\bx$ and adversarial examples $\bxa$ generated attacking with GCG the \texttt{Bert-like (grad. direction)} model.}
\vspace{0.5cm}
\end{table}

\begin{table}[htbp]
\centering
\footnotesize
\begin{tabularx}{\linewidth}{lXXc}\toprule
 & \textbf{Input} & \textbf{Output $\by$} & \textbf{Loss} \\
\midrule
$\bx$~: & Ben and Sam were friends who liked to play with toys. They had many toys, but their favorite was a & \multirow{3}{*}{ helicopter. The helicopter }  & 13.76 \\
$\bxa$: & p nodd nodd clotZub fam Do un said,ball wr." hitac bi'reide.-x & &  \textbf{ 11.14 } \\
\midrule
$\bx$~: & Lily was excited to go to school. She liked to learn new things from her teach. She also & \multirow{3}{*}{ liked to play with her friends. Today, they }  & 17.10 \\
$\bxa$: & its botindbbougower. seenhere eezick moral food Gra peop everyoun'oun ever & &  \textbf{ 16.34 } \\
\bottomrule
\end{tabularx}
\caption{Original inputs $\bx$ and adversarial examples $\bxa$ generated attacking with GCG the \texttt{Inv-First (grad. direction)} model.}
\vspace{0.5cm}
\end{table}

\end{document}